\documentclass[letterpaper, 10 pt, conference]{ieeeconf}  

\IEEEoverridecommandlockouts                              

\usepackage{cite}
\usepackage{amsmath,amssymb,amsfonts}
\usepackage{algorithmic}
\usepackage{graphicx}
\usepackage{textcomp}
\usepackage{xcolor}
\usepackage{color, colortbl}
\usepackage{multirow}
\usepackage{siunitx}
\usepackage{subcaption}
\usepackage{float}
\usepackage{stfloats}
\usepackage{hyperref}
\usepackage{fancyhdr}
\usepackage{booktabs}
\usepackage{microtype}
\usepackage{booktabs}
\hypersetup{
    colorlinks=true,
    linkcolor=black,
    filecolor=magenta,      
    urlcolor=cyan,
    pdftitle={DenoiseCP-Net},
    pdfpagemode=FullScreen,
    }

\title{\LARGE \bf
DenoiseCP-Net: Efficient Collective Perception in Adverse Weather via Joint LiDAR-Based 3D Object Detection and Denoising
}

\author{Sven Teufel$^{1}$, Dominique Mayer$^{1}$,  Jörg Gamerdinger$^{1}$ and Oliver Bringmann$^{1}$
\thanks{$^{1}$University of Tübingen, Faculty of Science, Department of Computer Science, Embedded Systems {\tt\small \{sven.teufel, dominique.mayer joerg.gamerdinger, oliver.bringmann\} @uni-tuebingen.de}}%
}

\begin{document}

\maketitle
\thispagestyle{empty}
\pagestyle{empty}

\begin{abstract}
While automated vehicles hold the potential to significantly reduce traffic accidents, their perception systems remain vulnerable to sensor degradation caused by adverse weather and environmental occlusions. Collective perception, which enables vehicles to share information, offers a promising approach to overcoming these limitations. However, to this date collective perception in adverse weather is mostly unstudied. Therefore, we conduct the first study of LiDAR-based collective perception under diverse weather conditions and present a novel multi-task architecture for LiDAR-based collective perception under adverse weather. Adverse weather conditions can not only degrade perception capabilities, but also negatively affect bandwidth requirements and latency due to the introduced noise that is also transmitted and processed. Denoising prior to communication can effectively mitigate these issues. Therefore, we propose DenoiseCP-Net, a novel multi-task architecture for LiDAR-based collective perception under adverse weather conditions. DenoiseCP-Net integrates voxel-level noise filtering and object detection into a unified sparse convolution backbone, eliminating redundant computations associated with two-stage pipelines. This design not only reduces inference latency and computational cost but also minimizes communication overhead by removing non-informative noise. We extended the well-known OPV2V dataset by simulating rain, snow, and fog using our realistic weather simulation models. We demonstrate that DenoiseCP-Net achieves near-perfect denoising accuracy in adverse weather, reduces the bandwidth requirements by up to \SI{23.6}{\percent} while maintaining the same detection accuracy and reducing the inference latency for cooperative vehicles.
\end{abstract}

\section{Introduction}
\label{sec:intro}

The most likely cause of death for young people aged 5 to 29 are road traffic injuries. Especially in adverse weather such as rain and snow, the crash and injury rate can increase by \SIrange{70}{80}{\percent}~\cite{Death}. Automated vehicles aim to reduce the number of accidents. In order to achieve this, a comprehensive and correct perception of the environment is required; however, adverse weather or occlusions degrade the sensing capabilities of vehicle-local perception~\cite{teufel2023enhancing, volk2019towards}. Collective Perception (CP) using multiple distributed vehicles is a promising approach to overcome the limitations of vehicle-local perception. A major challenge in CP is the exchange and fusion of information between different agents. CP methods can be categorized into three classes based on the shared information. Late fusion approaches use preprocessed data and exchange object states, which results in a loss of information. Early fusion uses raw sensor data for information exchange, which exceeds current bandwidth limitations and therefore can be considered as unsuitable in practice. The intermediate fusion uses features from neural networks as information to exchange, this meets the communication channel resources and reduced the loss of information due to the pre-processing; however, for intermediate fusion all cooperative agents must use the same backbone since it otherwise suffers from a feature domain gap, this can be considered as unpractical for real-world application. For LiDAR-based collective perception, the utilization of sparse voxel grids as environment representation as proposed in ~\cite{teufel2024dynamic} allows to efficiently exchange spatial information without a significant loss in detail by discretizing point clouds into voxel grids. In order to reduce the bandwidth requirement, only the coordinates of voxels that contain points are transmitted, omitting any additional features. Unfortunately, adverse weather such as rain, snow, and fog leads to noise in the point cloud ~\cite{teufel2022simulating}. This noise can not only negatively affect the perceptual performance but also can lead to a higher number of voxels which then increases the communication load with unnecessary data being transmitted. Therefore, we propose a novel denoising and fusion architecture for LiDAR-based collective perception, called \textbf{DenoiseCP-Net}. DenoiseCP-Net jointly fuses the collectively shared sparse voxel grids with the local sensor data and simultaneously denoises the sparse voxel grid created from the ego LiDAR sensor. 
DenoiseCP-Net uses a shared backbone for both tasks, saving computation and reducing latency compared to a two-stage approach.\\
Our main contributions are:
\begin{itemize}
\setlength\itemsep{0.5em}
    \item We conduct the first study on the impact of rain, snow and fog on LiDAR-based collective perception.
    
    \item We present a novel and effective multi-task architecture for sparse voxel grid fusion in collective perception and denoising of sparse voxel grids called DenoiseCP-Net.
    
    \item DenoiseCP-Net achieves strong robustness against adverse weather conditions while simultaneously achieving high denoising performance.
    
    \item DenoiseCP-Net can significantly reduce the communication load in adverse weather.
\end{itemize}

\noindent In Sec.~\ref{rel_work} we present related work to our approach. Afterwards, we introduce the utilized environment representation for collective perception in Sec.~\ref{sec:representation}. In Sec.~\ref{sec:Weather_simulation} we present the weather simulation models used to augment the dataset with realistic weather conditions. Then we introduce the DenoiseCP-Net architecture in Sec.~\ref{sec:DenoiseCP}. After that, Sec~\ref{sec:eval} describes the conducted experiments including the augmented datasets. The results are provided in Sec.~\ref{sec:results}. Finally, we give a conclusion and outlook on future research.

\section{Related Work}
\label{rel_work}
\subsection{Collective Perception Methods}
Collective perception enhances object detection by enabling multiple agents to share data of the environment. Fusion strategies are typically categorized into early, intermediate, and late fusion. These categories each come with a specific trade-off regarding performance and bandwidth. Early fusion approaches rely on the sharing of raw sensor data that is fused with the local sensor data before being processed. Chen et al. \cite{chen2019cooper} showed a significant performance improvement using early fusion compared to local only detection, however the datasets used for evaluation are not suited for this task. Xu et al. \cite{xu2022opv2v} benchmarked several early fusion methods on their dataset. The results showed that early fusion methods achieve a higher performance than late fusion methods. However, the bandwidth requirement is too high to be used in practice. The intermediate fusion approach relies on the exchange of neural network features. Chen et al. \cite{chen2019f} extended their early fusion approach by sharing voxel or spatial features to reduce the transmitted data. Sharing voxel features performed similar to the raw data sharing while reducing the transmitted data slightly. Similar to the sharing of voxel features, Bai et al. \cite{bai2021pillargrid} shared pillar features from the PointPillars \cite{lang2019pointpillars} detector. Similar to Chen et al.~\cite{chen2019f}, their results show a significant improvement in detection performance compared to local-only detection. The sharing of keypoint features of the PV-RCNN detector was evaluated by Yuan et al. \cite{yuan2022keypoints}. They could show that sharing keypoint features can significantly reduce the amount of shared data compared to sharing entire feature maps, while achieving better detection performance. All these approaches were able to show a strong detection performance while reducing the amount of data transmitted, however, they all suffer from a domain gap when different neural architecture are used to generate the features \cite{xu2023bridging}. For late fusion, detected bounding boxes are exchanged between vehicles. A simple way to realize late fusion is to use detected bounding boxes from multiple cooperative agents and weight them based on their detection confidence which is used in multiple approaches for local sensor fusion and collective perception ~\cite{solovyev2021weighted, houenou_track--track_2012, muller_generic_2011, aeberhard_object-level_2017}. Another way of late fusion is to use an adapted Kalman filter. Approaches using Kalman filter based collective perception are presented in~\cite{allig2019,volk_environment-aware_2019,gabb2019infrastructure,volk2021}. 
Another late fusion method is to directly fuse the collective detections within the local perception pipeline using neural networks \cite{teufel2023collective}. While Late fusion has the lowest requirement regarding the bandwidth and standards already exist, it has usually the lowest performance induced by the information loss in the pre-processing. Besides these common fusion approaches Teufel et al. \cite{teufel2024dynamic} presented MR3D-Net which relies on the sharing of sparse voxel grids constructed from the LiDAR point clouds. MR3D-Net outperformed all early fusion methods benchmarked by Xu et al. \cite{xu2022opv2v} while reducing the bandwidth requirement by up to \SI{94}{\percent}, making it feasible to use in practice.

\subsection{Collective Perception in Adverse Weather}
Collective perception in adverse weather is mostly unaddressed in research, primarily caused by the lack of appropriate collective perception datasets containing adverse weather scenarios \cite{teufel2024collective}. However, there are recent works addressing the robustness of LiDAR-based collective perception in rain and fusion of LiDAR and RADAR in fog and snow.

Jiang et al. \cite{jiang2024weather} proposed a framework called Co-Denoising to improve collective perception under rain conditions. Their method incorporates a two-stage denoising process with a heuristic sampling-based approach at the local level to filter noisy data and a bayesian neural network-based refinement at the global aggregation stage. The authors claim that their approach significantly improves robustness in rainy conditions, but their reported results on the OPV2V datasets raise serious concerns, as they are unrealistically high and the reported results of other state-of-the-art methods differ drastically from the official results provided by the OPV2V benchmark, making their entire evaluation of the proposed method highly questionable.

Huang et al. \cite{huang2024v2x} introduced V2X-R, a collective perception dataset that integrates LiDAR and 4D radar. Within their LiDAR-4D radar fusion pipeline, they proposed a Multi-modal Denoising Diffusion module designed to address challenges posed by adverse weather conditions. This module leverages radar features, which are robust to environmental disturbances, as conditions to guide the diffusion process and denoise noisy LiDAR features, enhancing the overall perception performance in challenging weather conditions.  

\begin{figure*}
    \centering
    \begin{subfigure}[t]{0.329\textwidth}
        \includegraphics[width=\textwidth, trim = 8cm 0cm 8cm 0cm, clip]{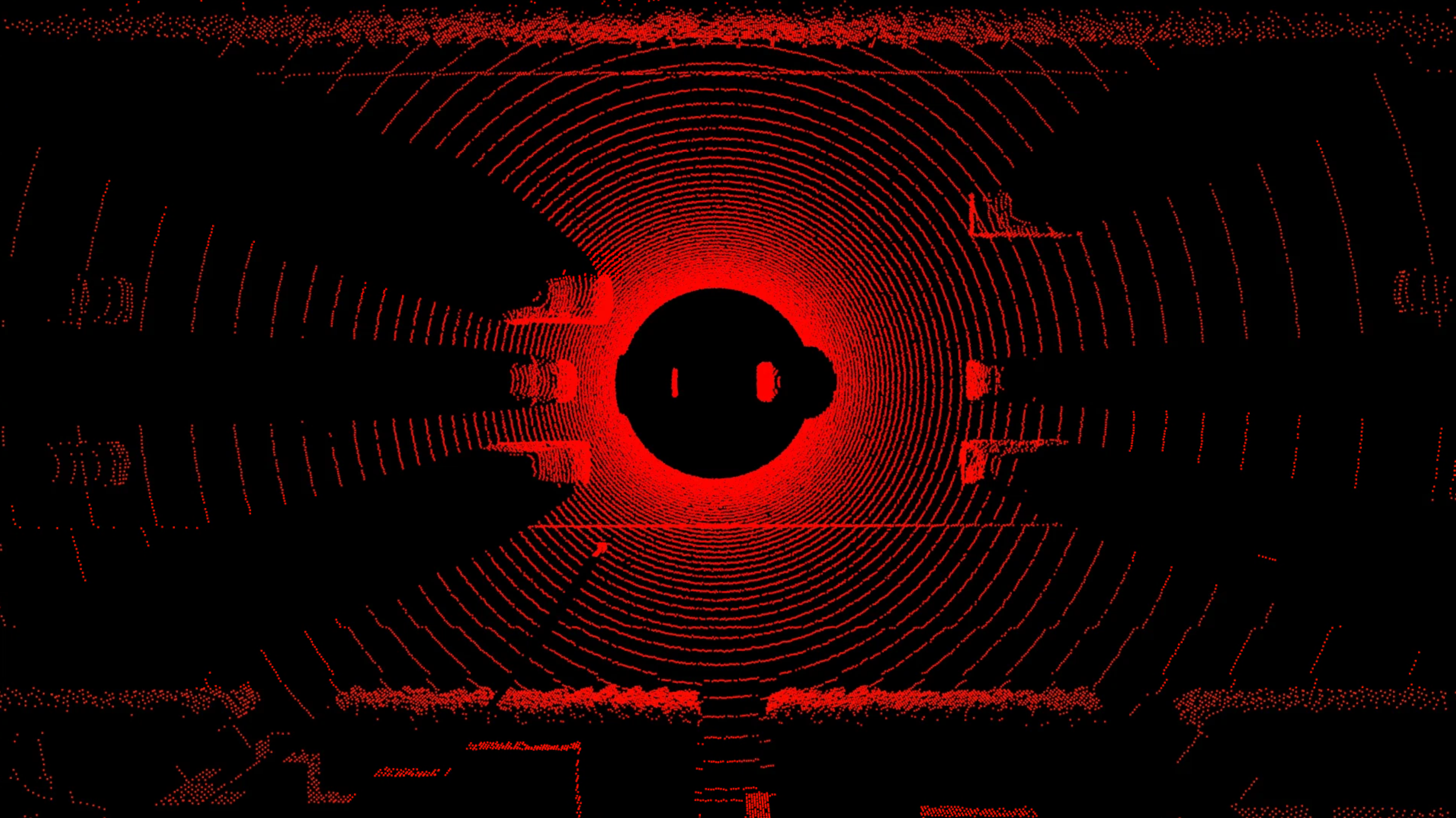}
        \caption{Original Point Cloud}
        \label{fig:sub1}
    \end{subfigure}
    \begin{subfigure}[t]{0.329\textwidth}
        \includegraphics[width=\textwidth, trim = 8cm 0cm 8cm 0cm, clip]{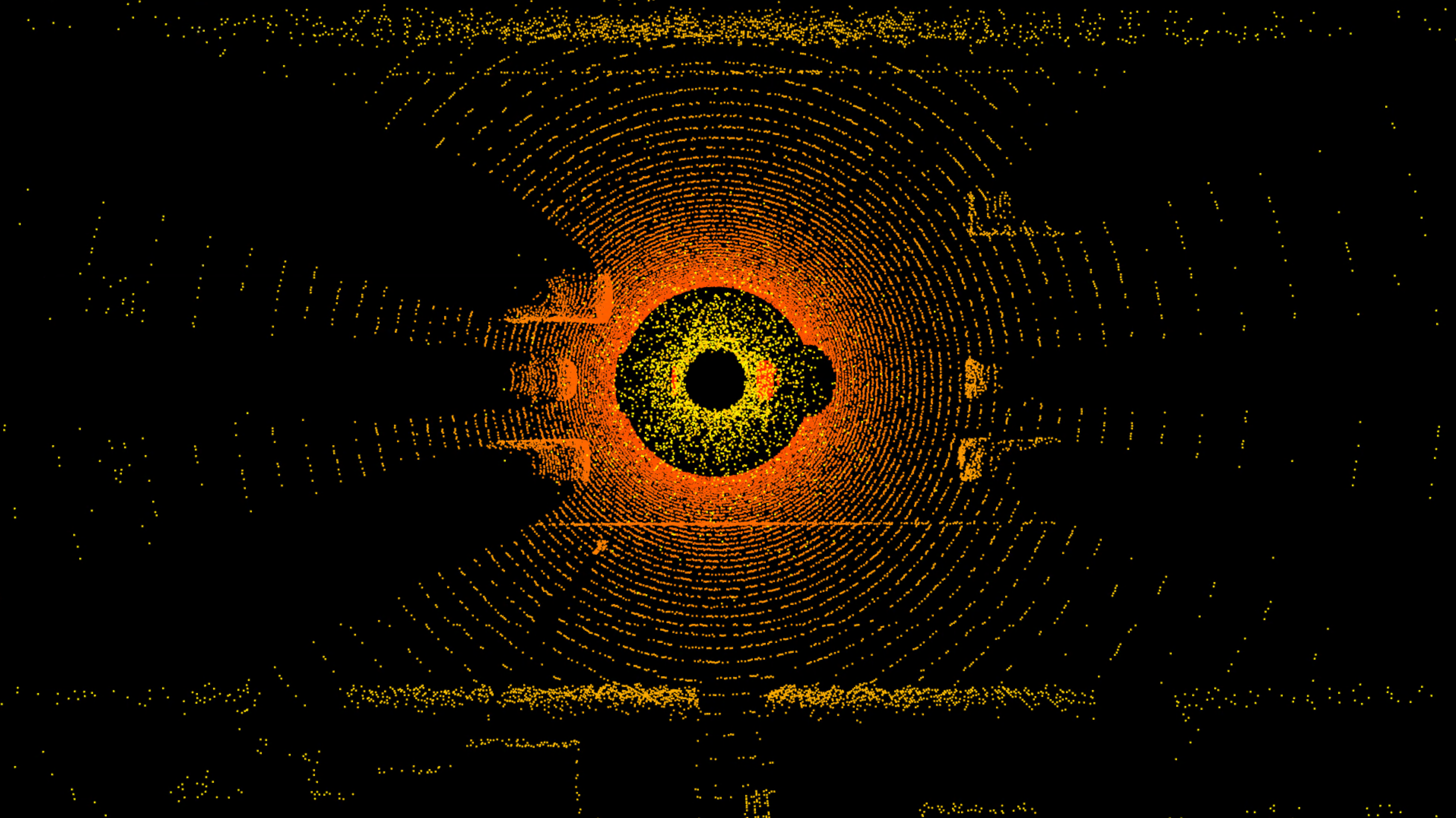}
        \caption{Simulated Fog}
        \label{fig:sub2}
    \end{subfigure}
    \begin{subfigure}[t]{0.329\textwidth}
        \includegraphics[width=\textwidth, trim = 8cm 0cm 8cm 0cm, clip]{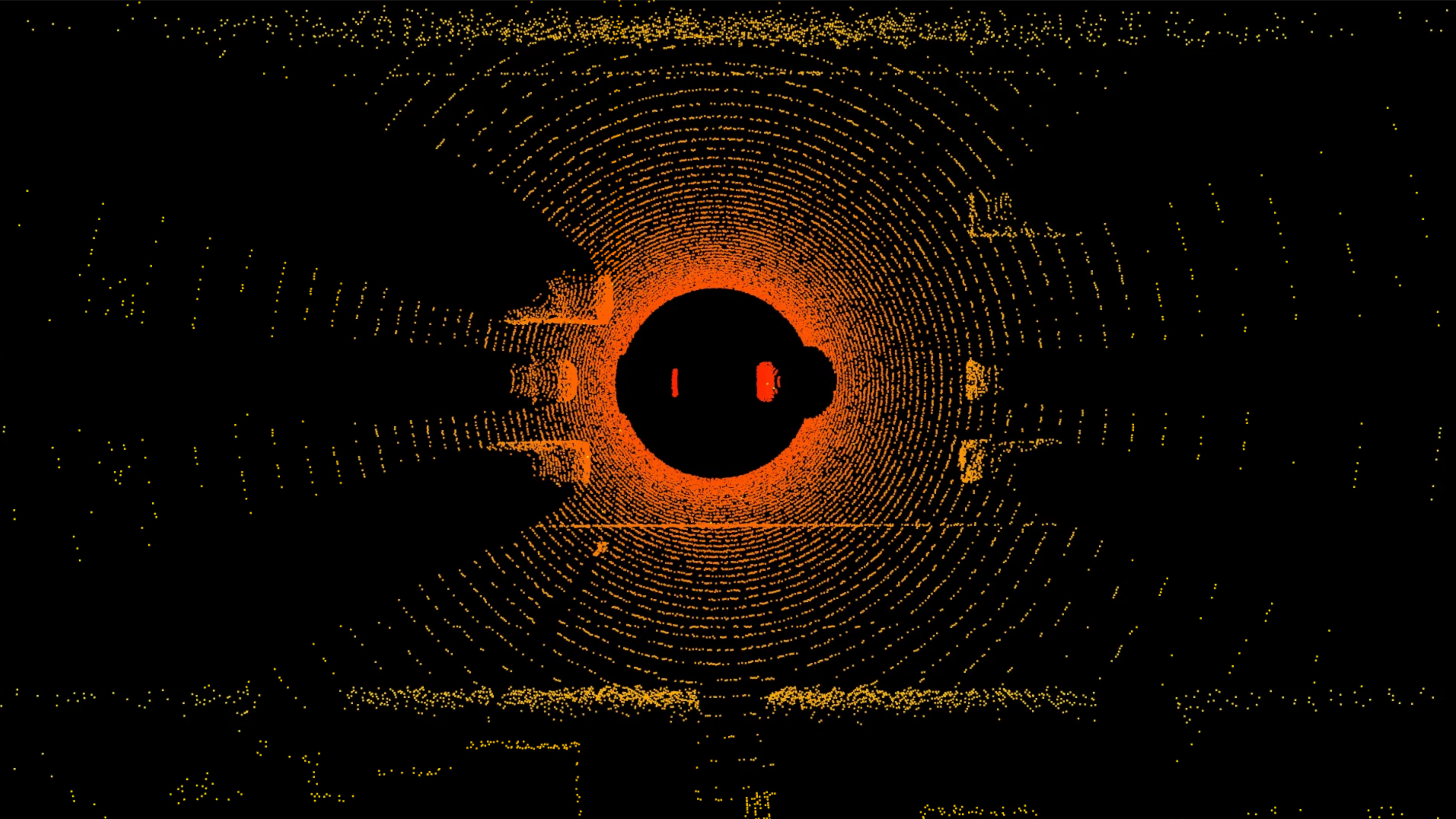}
        \caption{Denoised Point Cloud}
        \label{fig:sub3}
    \end{subfigure}
    \caption{LiDAR point clouds from the OPV2V \cite{xu2022opv2v} in bird's-eye view. (a) Original point cloud, (b) the point cloud with simulated fog using the simulation model from Sec. \ref{sec:Weather_simulation_fog}, (c) the foggy point cloud after denoising with DenoiseCP-Net}
    \label{fig:main}
\end{figure*}

\section{Environment Representation}
\label{sec:representation}

In contrast to other existing CP methods, DenoiseCP-Net utilizes sparse voxel grids as environment representation instead of raw sensor data, feature maps or object states. As shown by MR3D-Net \cite{teufel2024dynamic} sparse voxel grids pose several advantages compared to other environment representations used in collective perception. Sparse voxel grids are way more compact than raw LiDAR point clouds while still preserving a high level of detail. Furthermore, sparse voxel grids are more expressive than object states, they do not suffer from feature domain gaps as neural network feature maps, and sparse voxel grids can be directly processed by 3D convolutions. In DenoiseCP-Net, the sparse voxel grids do not contain any features, only the coordinates of the voxel grids are exchanged to reduce the amount of transmitted data. As input to the model, the center points of the voxels are used as features, which can be calculated from the coordinates, voxel size, and origin of the voxel grid. When dealing with the impact of adverse weather conditions, potential problems could arise when using sparse voxel grids as environment representation. First, due to the introduced noise by the weather conditions, the bandwidth requirement could rise. This is caused since a large amount of the data reduction when building sparse voxel grids from point clouds is caused by points only laying on a surface that is then represented by only a small number of voxels, however, when noise is distributed in free space, this leads to a significantly higher number of voxels. 

Another problem that might arise is that due to the omittance of all point features, including the intensity, it could be harder to differentiate between objects and noise, resulting in a performance drop of the fusion model. A potential solution to both these problems is to denoise the sparse voxel grid based on the point features before it is transmitted to other traffic participants. Therefore, we investigate the impact of adverse weather conditions on sparse voxel grid based collective perception in detail and study the benefits of sharing denoised sparse voxel grids.

\section{Weather Simulation}
\label{sec:Weather_simulation}
Since there is no real-world dateset for collective perception in adverse weather available, we use realistic weather simulation models for LiDAR point clouds in order to evaluate collective perception in the desired weather conditions. As simulation models we use the rain, snow and fog simulation for LiDAR point clouds from our previous work \cite{teufel2022simulating}. In this section we will discuss each simulation model briefly, for a more detailed description of the weather simulations see \cite{teufel2022simulating}.

\subsection{Rain Simulation}
The rain simulation is a physically based simulation model that uses a noise filter, which consists of uniformly distributes spherical particles within the sensors field of view to represent raindrops. Ray tracing is then used to approximate the attenuation effects caused by scattering and absorption that occur when LiDAR pulses interact with these particles. The sizes of the raindrops follow the Feingold-Levin \cite{feingold1986lognormal} log-normal distribution for raindrop sizes, capped at \SI{6}{\milli\metre}, representing the maximum stable diameter under typical conditions. This distribution reflects the frequency of different drop sizes in a rain event, based on empirical findings. 

To account for beam divergence, which describes the spreading of the laser beam of the LiDAR as it travels through space, for each point in the point cloud multiple rays are generated within the divergence angle to generate a discretized cone, which covers the beams cross section evenly.

Then for each ray it is determined whether it intersects with raindrops within the noise filter. To approximate the total portion of light that is attenuated, a hit ratio, representing the proportion of rays intersecting with any raindrop, is calculated. Based on this hit ratio, the points from the point cloud are selected for modification. If there is a raindrop that back-scatters enough light to cause a false measurement, i.e. the ratio of rays intersecting this raindrop and the number of intersecting rays exceeds a threshold, the location of this raindrop is used as the new location for this point. Otherwise, the point is deleted. Finally, the point intensities are adjusted based on empirical findings. This simulation model can be parametrized by the precipitation rate and the amount of raindrops
per volume.

\subsection{Snow Simulation}
The snow simulation is based on the same ray tracing approach as the rain simulation, but is specifically adapted to account for the unique physical properties of snowflakes. Unlike raindrops, snowflakes are typically larger, more irregular in shape, and composed of ice crystals or aggregates, which lead to distinct scattering and absorption effects. The size distribution of snowflakes is modeled using the Gunn-Marshall \cite{gunn1958distribution} distribution, optimized by Sekhon and Srivastava \cite{sekhon1970snow}. Since the distribution represents the molten diameters of snowflakes, a scaling factor ranging from 1 to 5 is applied to obtain realistic snowflake sizes. Smaller scaling factors represent wet, compact snow, while larger factors account for dry dendritic aggregates. In order to account for the snowflakes characteristics, also the thresholds of the rain model are adjusted to better represent realistic snowfall. Like in the rain model, the point intensities are also adjusted based on empirical findings. This simulation model is parametrized by the precipitation rate, the amount of snowflakes, and the scaling factor.

\subsection{Fog Simulation}
\label{sec:Weather_simulation_fog}
Unlike rain and snow, where the droplet concentration is relatively low, fog contains an extremely high number of droplets per unit volume. Due to this vast droplet density, a ray tracing approach to simulate fog effects on LiDAR point clouds is computationally infeasible. Instead, the fog simulation employs an efficient probabilistic model that assumes evenly dense fog. This model generates a continuous noise field within the point cloud, effectively capturing the scattering and attenuation effects of fog without the computational overhead of ray tracing. The probabilistic point selection and modification process is based on the physical properties of the attenuation effects in fog and was fitted to real fog chamber recordings under controlled conditions in order to resemble the characteristic of real foggy conditions. The fog simulation is parametrized by the maximum viewing distance.

\section{DenoiseCP-Net}
\label{sec:DenoiseCP}

\begin{figure*}
    \centering
    \includegraphics[trim= 1.6cm 3cm 1.3cm 3.3cm, clip, width=\textwidth]{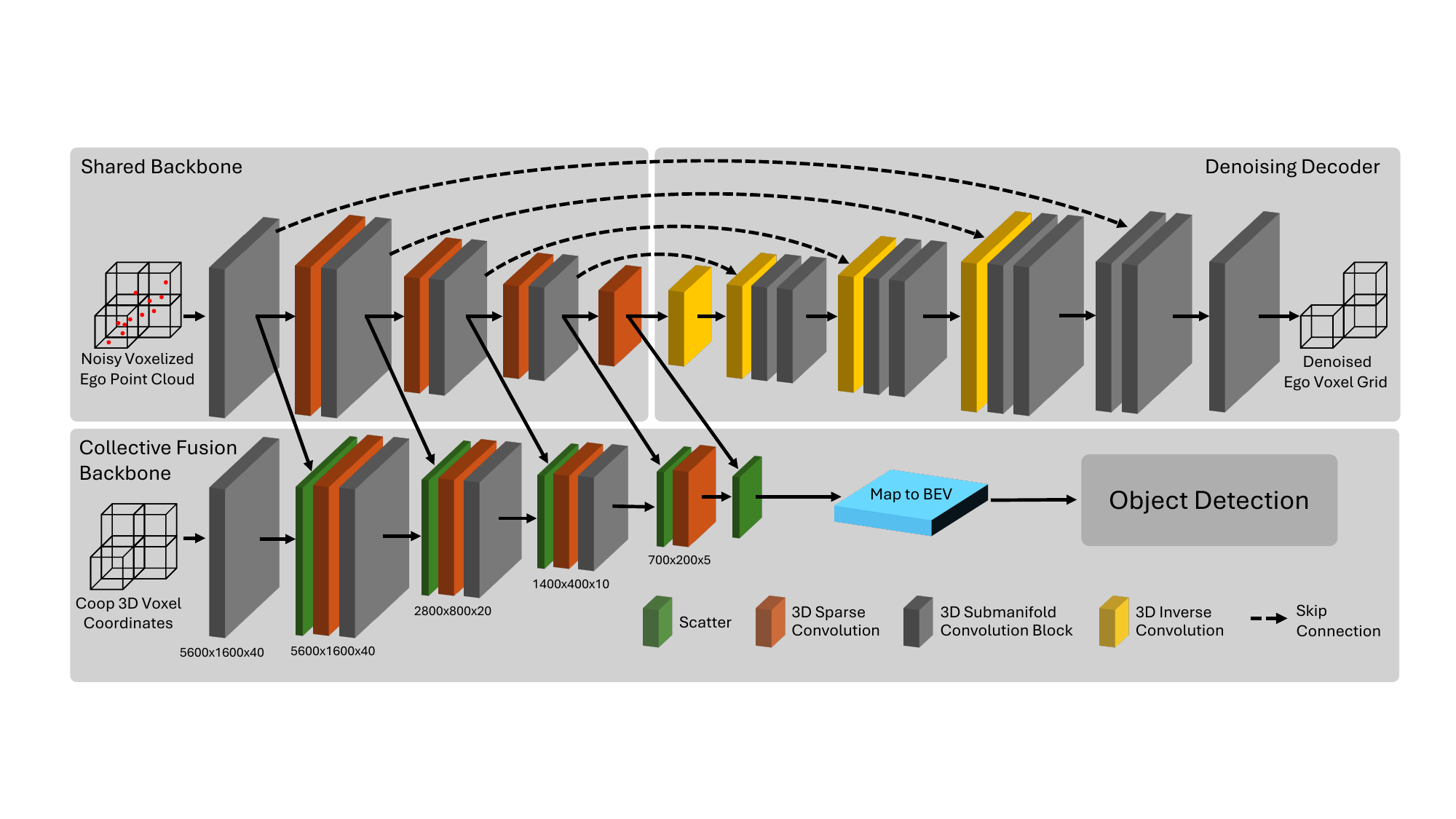}
    \caption{Architecture of DenoiseCP-Net. The shared backbone takes the noisy ego voxel grid as input and extracts high dimensional features that are used both for denoising and object detection. The collective fusion backbone takes the received sparse voxel grids of cooperative vehicles as input and fuses them with the ego feature. The output feature map of the collective fusion backbone is then mapped to BEV and used for object detection. The denoising decoder upsamples the sparse voxel grids to the original spatial resolution, following a U-Net architecture and then predicts class scores for each voxel.}
    \label{fig:architecture_overview}
\end{figure*}

Real‑time autonomous driving demands precise object detection with minimal data overhead. Rain, snow, and fog introduce noise into LiDAR point clouds, which not only can increase computation and latency due to an increased amount of data \cite{li2020deep}, but also degrades detection accuracy \cite{teufel2022simulating}. Since the noise contains no valuable information for the underlying task, filtering out this noise can reduce the bandwidth requirement for the communication and the processing demand for cooperative vehicles. To leverage this insight, we introduce \textbf{DenoiseCP-Net}, a multi‑task model that combines 3D object detection with voxel‑level noise removal. By using a shared sparse‑convolutional backbone for both detection and denoising, DenoiseCP-Net lowers computational demands compared to a two-stage pipeline with separate backbones and reduces data transmission by removing noisy voxels before sharing. An overview of the proposed DenoiseCP-Net architecture is given in Fig. \ref{fig:architecture_overview}.
\subsection{Architecture}
DenoiseCP-Net consists of three main parts, the \textit{shared backbone}, the \textit{collective fusion backbone} and the \textit{denoising decoder}.
Inspired by S2S-Net \cite{teufel2025s2s} the \textit{shared backbone} and \textit{collective fusion backbone} are two parallel, structurally identical sparse-convolutional backbones for feature extraction from different sources. The \textit{shared backbone} processes the ego vehicle’s noisy voxel grid, and the \textit{collective fusion backbone} processes voxel grids received from other cooperative vehicles. Both backbones start with an initial submanifold convolution block, followed by three downsampling stages, each combining a strided 3D sparse convolution with a submanifold block. Each submanifold convolution block consists of two submanifold convolution layers. Finally, another sparse convolution is applied to produce bottleneck features. 

The bottleneck features of the shared backbone are then fed into a U-Net–style denoising decoder that mirrors the encoder: at each level, the decoder concatenates its feature map with the matching encoder skip features, upsamples via a 3D inverse convolution, and refines the result with two 3D submanifold convolution blocks. A last submanifold convolution then produces per-voxel two-class outputs. The resulting voxel grid is identical to the ego input voxel grid with \textit{noise}/\textit{no noise} class predictions for each voxel. All voxels which are predicted as \textit{noise} are then deleted before the sparse voxel grid is shared with the other cooperative vehicles.

For the 3D object detection to integrate both the local and collective information, the scatter operation is applied before every encoding stage in the \textit{collective fusion backbone} to incorporate the features from the shared backbone into the collective fusion backbone. The resulting feature map is then mapped to bird's-eye view and used as input to an object detector, in our case PV-RCNN++ \cite{shi2023pv}. 

\subsubsection{Scatter Operation}
In order to fuse the features from the shared backbone with the features from the collective fusion backbone, we apply the scatter operation, since a feature channel concatenation would either result in an imbalance in the feature dimension or massive padding would be necessary which is computationally expensive. In DenoiseCP-Net, the scatter operation employs an element-wise maximum function on the feature vectors of voxels present in both the voxel grid of the shared backbone and the collective fusion backbone. This results in a single feature vector for these voxels. The remaining voxels, which only appear in either one of the voxel grids, are then concatenated with the scattered voxels. This process leads to a fused voxel grid, which possesses a uniform number of channels for all voxels. As shown in Fig. \ref{fig:architecture_overview} the scatter operation is applied in the collective fusion backbone before each sparse convolution block, except the first one since the feature channels differ in the input, to incorporate the shared backbone features with the ego sensor data. 

\subsubsection{Implementation Details}
For both the shared and collective backbone, we use a voxel size of $5\times5\times10\,$cm and a maximum grid extent of $280\times80\times4\,$m, yielding input dimensions of $5600\times1600\times40$ voxels. In each backbone, the 2nd, 3rd and 4th sparse‐convolution layers employ a stride of 2 to halve spatial resolution. All sparse‐convolution blocks employ a $3\times3\times3$ kernel and channel widths of 16, 32, 64, and 64. 

The denoising decoder is based on a sparse U-Net approach, featuring four upsampling stages followed by a refinement block. Starting at the shared backbone’s bottleneck, a sparse inverse convolution with stride 2 doubles spatial resolution and produces 64 channels. Three additional inverse convolutions then restore full resolution while reducing channel count to 64, 32, and 16 respectively. Before each upsampling stage, the decoder merges its feature map with the corresponding encoder skip connection by concatenating voxel features and refines the combined sparse voxel grid with two residual 3D submanifold convolution blocks. A final stage, without any further upsampling, consists of two submanifold convolution blocks that further refine the high-resolution features. In total, four skip connections preserve fine spatial detail, and a concluding $1 \times 1 \times 1$ submanifold convolution maps each voxel to the final two-class output. 

\section{Evaluation}
\label{sec:eval}

\subsection{Datasets}
\label{sec:datasets}
For evaluation, we use the OPV2V dataset \cite{xu2022opv2v}, which comprises 73 distinct urban and suburban scenarios, split into 6\,765 training and 2\,170 test samples. 
One sample corresponds to one time step with multiple recorded frames of different vehicles. The ego vehicle is randomly selected from the present vehicles for each sample, the other frames are then used as the collectively shared information. OPV2V contains 2 to 7 cooperative vehicles equipped with sensors and additional non-cooperative vehicles as traffic. 

With the weather simulation models explained in Sec.~\ref{sec:Weather_simulation}, we generated four weather-augmented variants of the entire dataset, namely Snow, Rain, Light Fog and Dense Fog, by randomly sampling one set of simulation parameters per scenario and applying those fixed values across all time steps and vehicles within that scenario. Each scenario's parameters remain constant throughout its sequence, guaranteeing uniform conditions for all frames and agents. We use the following parameter sampling ranges:

\begin{table}[H]
\normalsize
\centering
\caption{Weather simulation parameter ranges}
\label{tab:weather-params}
\begin{tabular}{@{}l|l|l@{}}
\toprule
\textbf{Weather} & \textbf{Parameter} & \textbf{Range} \\ 
\midrule
\multirow{3}{*}{Snow} 
& Precipitation rate & \SIrange{5}{20}{\milli\metre\per\hour} \\
\addlinespace[1pt]
& Snowflake density & \SIrange[per-mode=power]{500}{2000}{\per\cubic\metre} \\
\addlinespace[1pt]
& Scaling factor & \SIrange{2}{5}{} \\ 
\midrule
\multirow{2}{*}{Rain} 
& Precipitation rate & \SIrange{20}{50}{\milli\metre\per\hour} \\
\addlinespace[1pt]
& Raindrop density & \SIrange[per-mode=power]{1000}{2000}{\per\cubic\metre} \\ 
\midrule
Light Fog & Viewing distance & \SIrange{70}{200}{\metre} \\
\midrule
Dense Fog & Viewing distance & \SIrange{30}{100}{\metre} \\
\bottomrule
\end{tabular}
\end{table}

To cover a wide variety for each weather condition we choose these parameter ranges from low intensity with minor weather impact to a high intensity with strong weather effects. Since fog causes a drastic degradation of sensor data at low visibility distances, we choose to create two fog datasets, a light fog dataset where the very low visibility distances are excluded to resemble frequent occurring foggy conditions and a dense fog dataset where they are included to also incorporate rarely occurring extreme conditions. The false points caused from the rain, snow, or fog simulation are then labeled \textit{noise}, while all other points are labeled \textit{no noise}. During voxelization, these labels are propagated to voxels by majority vote to provide the voxel-level ground truth.

\vspace*{-2mm}
\subsection{Experiments}
We conducted one experiment for each combination of our four weather-augmented datasets Rain, Snow, Light Fog and Dense Fog and the three cooperative schemes Noisy, Denoised and Mixed. In each experiment we train the complete end-to-end pipeline together with the PV-RCNN++ object detector for 120 epochs, keeping all hyperparameters constant. During training we randomly select one vehicle as ego in each scene and process its noisy sparse voxel grid with the shared backbone, while the cooperative vehicles’ grids flow through the collective fusion backbone and merge via the scattering operation. In the Noisy scheme every agent transmits unfiltered voxel grids including the weather-induced noise, providing a baseline to quantify the raw impact of adverse conditions. By contrast, the Denoised scheme assumes all participants apply the denoising decoder to remove noise before sharing, representing ideal fully denoised cooperative conditions. The Mixed scheme models a heterogeneous deployment: for each scene half of the cooperative vehicles denoise their sparse voxel grids prior to transmission while the other half transmits noisy data. After training we evaluate both object detection performance and denoising metrics on the matching test split of each augmented dataset and record the average transmission bandwidth as well as the average inference latency. Furthermore we conducted a high intensity weather experiment for each weather condition, to quantify the impact of extreme weather conditions on the bandwidth and the possible reduction through denoising. 

\begin{table}
\normalsize
\centering
\caption{Denoising results for each weather condition}
\label{tab:denoising-results}
\renewcommand{\arraystretch}{1.2}
\begin{tabular}{@{}l|cc|cc@{}}
\toprule
\textbf{Weather} & \multicolumn{2}{c|}{\textbf{Accuracy [\%]}} & \multicolumn{2}{c}{\textbf{F1-Score [\%]}} \\
                 & \textit{noise} & \textit{no noise}          & \textit{noise} & \textit{no noise} \\
\midrule
Snow             & 99.73          & 99.94                      & 99.75          & 99.93 \\
\addlinespace[2pt]
Rain             & 99.77          & 99.99                      & 99.81          & 99.98 \\
\addlinespace[2pt]
Fog (Light)      & 99.98          & 99.99                      & 99.99          & 99.99 \\
\addlinespace[2pt]
Fog (Dense)      & 99.87          & 97.24                      & 96.03          & 98.58 \\
\bottomrule
\end{tabular}
\vspace*{-3mm}
\end{table}

\subsection{Evaluation Metrics}
\subsubsection{Object Detection}
For the evaluation, we report the Average Precision (AP) results on the test splits of our augmented datasets presented in Sec. \ref{sec:datasets} with an Intersection over Union (IoU) threshold of 0.7. As suggested by OPV2V \cite{xu2022opv2v}, we do not sort the detections by confidence for the AP calculation. We evaluate the detections within a range of $[-140, 140]$\,\si{\metre} in x-direction, $[-40, 40]$\,\si{\metre} in y-direction and $[-3, 1]$\,\si{\metre} in z-direction as this is the official evaluation range from OPV2V. 
Additionally, we limit the communication range to \SI{70}{\metre}, i.e. messages from the other cooperative vehicles, that are not within \SI{70}{\meter} range of the ego are ignored.
Furthermore, we calculate the required bandwidth under all conditions at a sensor frequency of \SI{10}{\hertz}. We only consider the actual data without any communication overhead. The reported bandwidth corresponds to the average bandwidth at which each vehicle sends data to the other vehicles.

\subsubsection{Denoising}
Since denoising condenses down to a binary classification task (\textit{noise} or \textit{no noise} for each voxel) we report standard classification metrics such as Accuracy and F1 score for the classes \textit{noise} and \textit{no noise}. Since there is a heavy class imbalance in some of the datasets between \textit{noise} and \textit{no noise}, we report all metrics per class. 

\subsubsection{Latency}
We furthermore measure the average inference latency of DenoiseCP-Net on the test split of the respective datasets. The latency is measured on a server with an AMD EPYC 9654P CPU and a NVIDIA H100 GPU using spconv 2.3.8 together with CUDA 11.8.

\section{Results}
\label{sec:results}

\begin{table*}
\centering
\caption{3D object detection results for cars, bandwidth reduction through denoising at a transmission frequency of \SI{10}{\hertz} and average inference latency on the OPV2V test split.}
\label{tab:bandwidth-reduction}
\renewcommand{\arraystretch}{1.2}
\setlength{\tabcolsep}{2pt}
\resizebox{\linewidth}{!}{%
\begin{tabular}{@{}l|ccc|ccc|cc@{}}
\toprule
\textbf{Weather}        & \multicolumn{3}{c|}{\textbf{Noisy}}                                                                                   & \multicolumn{3}{c|}{\textbf{Denoised}}                                                          & \multicolumn{2}{c}{\textbf{Reduction}}  \\
                        & AP@IoU$_{0.7}$                      & Bandwidth [\si{\mega\bit\per\second}]  & Inference         [\si{\milli\second}] & AP@IoU$_{0.7}$ & Bandwidth [\si{\mega\bit\per\second}] & Inference  [\si{\milli\second}]        & Bandwidth                & Inference    \\
\midrule 
Clear                    & 83.05                               & 14.40                                  & 55.7                                   & -          & -                                 & -                                  & -                  & -       \\
\addlinespace[1.5pt]
Snow                    & 61.29                               & 14.93                                  & 67.4                                   & 61.29          & 11.41                                 & 63.1                                   & 23.6 \%                  & 6.2 \%       \\
\addlinespace[1.5pt]
Rain                    & 80.82                               & 14.33                                  & 55.8                                   & 80.86          & 13.26                                 & 54.7                                   & 7.5 \%                   & 1.9 \%       \\
\addlinespace[1.5pt]
Fog (light)             & 66.85                            & 11.37                                  & 53.4                                   & 67.12          & 11.01                                 & 52.5                                   & 3.2 \%                   & 1.6 \%       \\
\addlinespace[1.5pt]
Fog (dense)             & 33.98                               & 8.61                                   & 56.2                                   & 33.11          & 6.64                                  & 53.9                                   & 22.8 \%                  & 4.0 \%       \\
\bottomrule
\end{tabular}}
\end{table*}

\begin{table}
\small
\centering
\caption{3D object detection performance for cars with mixed noisy and denoised shared data.}
\label{tab:mixed}
\renewcommand{\arraystretch}{1.2}
\setlength{\tabcolsep}{2pt}
\begin{tabular}{@{}l|c}
\toprule
\textbf{Weather}                     & AP@IoU$_{0.7}$                      \\
\midrule 
Snow                    & 57.59                                                                \\
\addlinespace[1.5pt]
Rain                    & 80.34                                                                \\
\addlinespace[1.5pt]
Fog (light)             & 67.13                                                                 \\
\addlinespace[1.5pt]
Fog (dense)             & 33.12                                                                 \\
\bottomrule
\end{tabular}
\vspace*{-3mm}
\end{table}

\subsection{Denoising}
In Tab.~\ref{tab:denoising-results} an overview of the results for the denoising task is provided. For all weather conditions very high results were achieved.
For snow an accuracy of \SI{99.73}{\percent} and \SI{99.94}{\percent} for \textit{noise} and \textit{no noise} could be achieved, similar results can be observed for the F1-Score with \SI{99.75}{\percent} and \SI{99.93}{\percent} for \textit{noise} and \textit{no noise}. For rain, the denoising performs slightly better than for snow and achieves an accuracy of \SI{99.77}{\percent} (\textit{noise}) and \SI{99.99}{\percent} (\textit{no noise}). Also the F1-Score increases by 0.06 p.p. and 0.05 p.p. for \textit{noise} and \textit{no noise} respectively. The best results for the denoising task was achieved for light fog. Here, the classification is nearly perfect with an accuracy of \SI{99.98}{\percent} and \SI{99.99}{\percent} for \textit{noise} and \textit{no noise}. The F1-score shows similar results with \SI{99.99}{\percent} for both classes. Compared to light fog, the denoising performance for the dense fog dataset is lower. While the noise accuracy still is \SI{99.87}{\percent}, the accuracy for no noise decreases by about 2.7 p.p. compared to the other weather conditions. Also, for the F1-Score a decrease can be observed; however, here the decrease is more significant for the noise class with nearly 4~p.p.. Despite this decrease, the F1-Score is still \SI{96.03}{\percent} for the \textit{noise} class. For the \textit{no noise} class the F1-Score reaches \SI{98.58}{\percent} which corresponds to a decrease of about 1.4 p.p. compared to the other weather conditions. 

These results show that denoising in adverse weather can be conducted very accurately, this is due to the noise points showing a different characteristics in their position, density, and intensity. Although being near perfect, these results are comparable to results from the literature on both real-world and simulated data~\cite{heinzler2020cnn, zhao2024triplemixer}.
Due to this near-perfect classification, the noisy voxels can be effectively removed from the sparse voxel grid that is shared with the other cooperative vehicles.

\subsection{Object Detection}
For the 3D object detection task, an overview of the results is given in Tab.~\ref{tab:bandwidth-reduction}. We provide results for object detection with noisy information from the other cooperative vehicles and with denoised information alongside with the required average bandwidth for the transmission of these information at \SI{10}{\hertz}. Additionally, we provide object detection results where a mix of both noisy and denoised data is transmitted, which is given in Tab. \ref{tab:mixed}.

On the original OPV2V datasets under clear conditions, DenoiseCP-Net achieved an AP of 83.05 with an average bandwidth of \SI{14.40}{\mega\bit\per\second}, this serves as a baseline for the evaluated weather conditions. A denoising was not performed for this dataset, since there is no noise in the original point clouds. 

For snow with noisy and denoised data an AP of 61.29 could be achieved, which corresponds to a performance drop of 21.71 p.p. compared to clear weather. With mixed data, a slightly lower performance was achieved with an AP of 57.59. The average bandwidth requirement increases compared to clear weather to \SI{14.93}{\mega\bit\per\second}, while with denial the bandwidth can be reduced to \SI{11.41}{\mega\bit\per\second}, which corresponds to a reduction of about \SI{23}{\percent}.
The lowest performance drop in adverse weather could be observed in rain with an AP of 80.82 for noisy and 80.86 for denoised data, this corresponds to a performance drop of 2.23 p.p. and 2.19 p.p.. As for snow, while noisy and denoised data show similar results, for the mixed data a slight decrease in performance can be observed, with an AP of 80.34. The bandwidth requirement for noisy data is \SI{14.33}{\mega\bit\per\second} on average, which is similar to clear weather and the bandwidth can be reduced by \SI{7.5}{\percent} with denoising.
For light fog, the performance is in between the performance in rain and snow with an AP of 66.85
for noisy data and 67.12 for the denoised data. For the mixed data in light fog a similar result with an AP of 67.13 was achieved.
Dense fog is the hardest condition in automotive perception with optical sensors, this can be also observed by the achieved results. In dense fog the perceptual performance in terms of AP decreases by about 49 p.p. compared to clear weather to an AP of 33.98 with noisy data and an AP of 33.11 for denoised data. As for the other weather conditions, the result with the mixed data is almost identical with an AP of 33.12.

For fog, the total number of points is significantly lower due to a stronger attenuation compared to rain and snow, this leads to a reduced bandwidth requirement of \SI{11.37}{\mega\bit\per\second} for light fog and \SI{8.61}{\mega\bit\per\second} for dense fog, respectively. Compared to clear conditions, this is already a reduction of \SI{21.0}{\percent} and \SI{40.2}{\percent}.
With denoising, the bandwidth can be reduced even further by \SI{3.2}{\percent} in light fog and \SI{22.8}{\percent} in dense fog, resulting in bandwidth requirements of \SI{11.01}{\mega\bit\per\second} and \SI{6.64}{\mega\bit\per\second}. 

These results are consistent with performance drops observed in real-world weather conditions \cite{teufel2023enhancing}.
For all weather conditions, the object detection performance is almost identical for noisy and denoised cooperative data, this shows that the removed noise does not contain any information that is valuable for the object detection task and therefore can be removed to lower the bandwidth. For all conditions, a significant reduction in the average bandwidth requirement could be achieved. 

\subsection{Latency}
For the latency reduction, similar results were achieved as for the bandwidth reduction, however, the percentage of reduction for the latency is lower than the reduction in bandwidth. This is caused by the fact that we measure the inference time of the whole DenoiseCP-Net architecture and the data reduction due to the denoising only affects the \textit{collective fusion backbone}, while the \textit{shared backbone} and the \textit{denoising decoder} are unaffected. The inference time with noisy data is similar to that in clear weather, except in snowy conditions, where a significantly increased inference time was observed. This is due to the increased number of voxels in snow, which increases the number of computations required. Like for the bandwidth, the largest reduction in inference latency due to denoising was observed in snowy conditions, with a reduction of \SI{6.2}{\percent}, followed by dense fog with \SI{4.0}{\percent}. For rain and light fog, only small latency reductions were observed with \SI{1.9}{\percent} and \SI{1.6}{\percent} respectively.

\subsection{High Intensity Weather Bandwidth}
The presented results for the bandwidth reduction on the augmented OPV2V dataset are averaged over the entire dataset. In high intensity weather conditions the possible reduction is significantly higher. In Tab.~\ref{tab:worst-case} the bandwidth results for the transmission of noisy and denoised sparse voxel grids under high intensity weather conditions is shown. For the simulation parameters of rain and snow, we selected the highest values from the ranges in Tab. \ref{tab:weather-params} and for the fog simulation we selected the lowest viewing distance. In these extreme conditions very high reduction rates are possible, in the case of fog and snow, the sparse voxel grids contain more noise than actual data, leading to a possible data reduction of \SI{52.3}{\percent} and \SI{62.3}{\percent} respectively. For the data in rain, the possible reduction is lower with \SI{8.5}{\percent}, this is due to a lower amount of introduced noisy, even at high intensities.

\begin{table}
\small
\centering
\caption{Bandwidth under high intensity weather conditions at a transmission frequency of \SI{10}{\hertz}.}
\label{tab:worst-case}
\renewcommand{\arraystretch}{1.2}
\begin{tabular}{@{}l|ccc@{}}
\toprule
\textbf{Weather} & \textbf{Noisy} & \textbf{Denoised} & \textbf{Reduction} \\
                  & [\si{\mega\bit\per\second}] & [\si{\mega\bit\per\second}] & [\si{\percent}] \\
\midrule
Fog    &  8.09 & 3.86  & 52.3 \\
\addlinespace[2pt]
Rain   & 14.97 & 13.69 &  8.5 \\
\addlinespace[2pt]
Snow   & 15.66 &  5.91 & 62.3 \\
\bottomrule
\end{tabular}
\vspace*{-3mm}
\end{table}

\section{Conclusion and Future Work}
\label{sec:conclusion}
In this work, we presented a novel multi-task architecture for LiDAR-based collective perception in adverse weather called DenoiseCP-Net. DenoiseCP-Net fuses collectively shared sparse voxel grids with the sparse voxel grids constructed from the ego sensor data for 3D object detection and simultaneously denoises the ego sparse voxel grid to reduce the communication bandwidth and to reduce the computational load and latency for the other cooperative vehicles. 

We demonstrated that DenoiseCP-Net achieves remarkable denoising performance under all evaluated conditions while showing strong robustness against adverse weather conditions for noisy and denoised collectively shared sparse voxel grids as well as for a mixture of both. The achieved denoising results are near perfect with greater \SI{97}{\percent} accuracy under all conditions, resulting in an efficient bandwidth reduction without sacrificing detection performance. The bandwidth requirement when sharing the denoised sparse voxel grids could be reduced by up to \SI{23.6}{\percent}. Due to the utilization of a shared backbone for both tasks the computational complexity as well as latency is reduced compared to a two-staged pipeline with separate backbones.

For future research, the presented DenoiseCP-Net architecture could be also utilized for further tasks such as joint semantic segmentation and 3D object detection, since the denoising task is structurally identical to semantic segmentation. Furthermore, this evaluation should be extended to real-world datasets for collective perception, if there will be a dataset available in the future.








\bibliographystyle{IEEEtran}
\bibliography{literature}

\end{document}